\documentclass[journal,12pt]{IEEEtran}
\usepackage{array,graphicx,epsfig,amssymb,amsfonts,bbm,amsmath,algorithm, algorithmic,epstopdf,subfigure,cite,tabularx}

\usepackage{latexsym}
\newtheorem{theorem}{\textbf{Theorem}}
\newtheorem{lemma}{\textbf{Lemma}}

\newenvironment{proof}[1][Proof]{\begin{trivlist}
\item[\hskip \labelsep {\bfseries #1}]}{\end{trivlist}}

\newcommand{\qed}{\nobreak \ifvmode \relax \else
      \ifdim\lastskip<1.5em \hskip-\lastskip
      \hskip1.5em plus0em minus0.5em \fi \nobreak
      \vrule height0.75em width0.5em depth0.25em\fi}

\begin{document}

\title{Fully Distributed and Asynchronized Stochastic Gradient Descent for Networked Systems}

\author{Ying Zhang \thanks{The author is with Department of Information Engineering, the Chinese University of Hong Kong.} }

\maketitle

\begin{abstract}
This paper considers a general data-fitting problem over a networked system, in which many computing nodes are connected by an undirected graph. This kind of problem can find many real-world applications and has been studied extensively in the literature. However, existing solutions either need a central controller for information sharing or requires slot synchronization among different nodes, which increases the difficulty of practical implementations, especially for a very large and heterogeneous system.

As a contrast, in this paper, we treat the data-fitting problem over the network as a stochastic programming problem with many constraints. By adapting the results in a recent paper \cite{wang2015random}, we design a fully distributed and asynchronized stochastic gradient descent (SGD) algorithm. We show that our algorithm can achieve global optimality and consensus asymptotically by only local computations and communications. Additionally, we provide a sharp lower bound for the convergence speed in the regular graph case. This result fits the intuition and provides guidance to design a `good' network topology to speed up the convergence. Also, the merit of our design is validated by experiments on both synthetic and real-world datasets.
\end{abstract}

\section{Introduction}
As one of the most important optimization algorithms, stochastic gradient descent (SGD) and many of its variants have been proposed to solve different optimization problems, and are gaining their popularity in this `big-data' era \cite{bottou2010large}. Popular examples include SVM, Logistic Regression for the convex cases \cite{bottou2012stochastic}, and deep neural networks for the non-convex cases \cite{dean2012large}. It can be shown that, for the basic version of SGD, the asymptotic convergence rate is $O(1/\sqrt{T})$ for the general convex problems and $O(1/T)$ for the strict convex problems, where $T$ is the number of iterations.

Nowadays, data is generated and collected at a higher speed due to the wide use of Internet, mobile devices, sensor networks, \textit{etc}. The basic version of SGD suffers from its nature of sequentially processing the data and is still too slow. To handle this issue, distributed versions of SGD have been tried by many practioneer \cite{dean2012large,noel2014dogwild} and analyzed by theoretical researchers \cite{hsieh2015passcode,recht2011hogwild,lian2015asynchronous}.

\begin{figure*}[tbh]
\centering
\subfigure[Server-Worker Structure]{\includegraphics[width=0.5\textwidth]{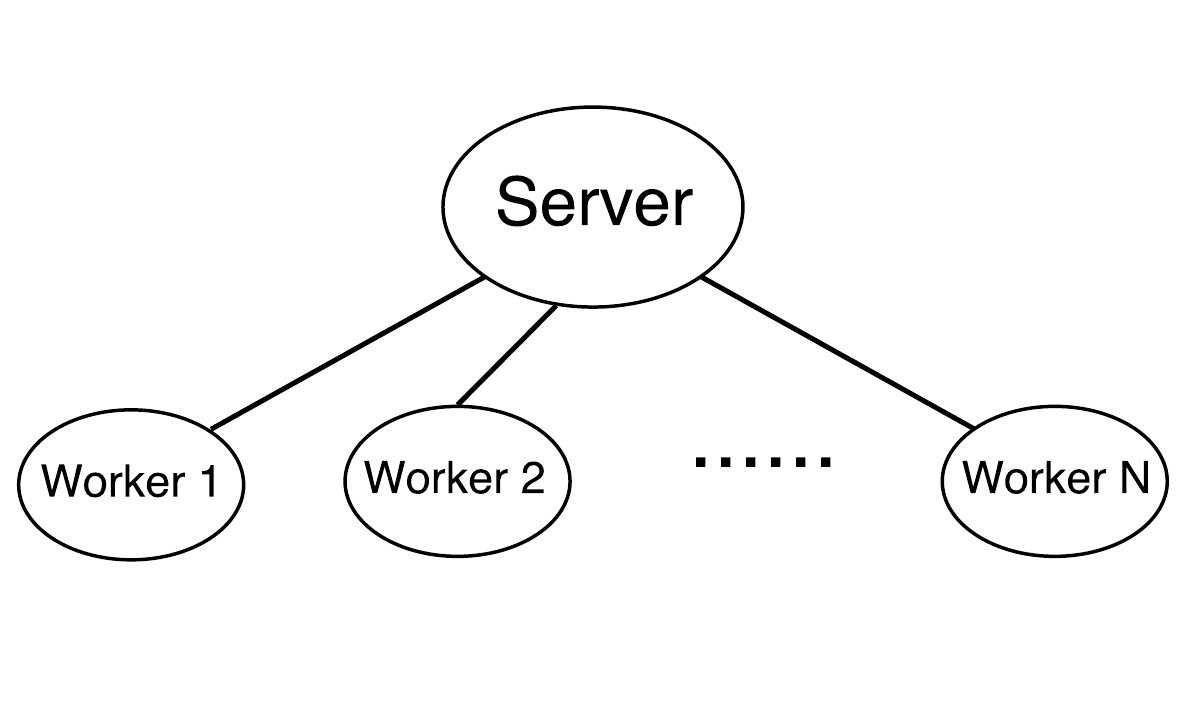}\label{fig:ServerWorker}}\hfill
\subfigure[Networked Structure (discussed in this paper)]{\includegraphics[width=0.4\textwidth]{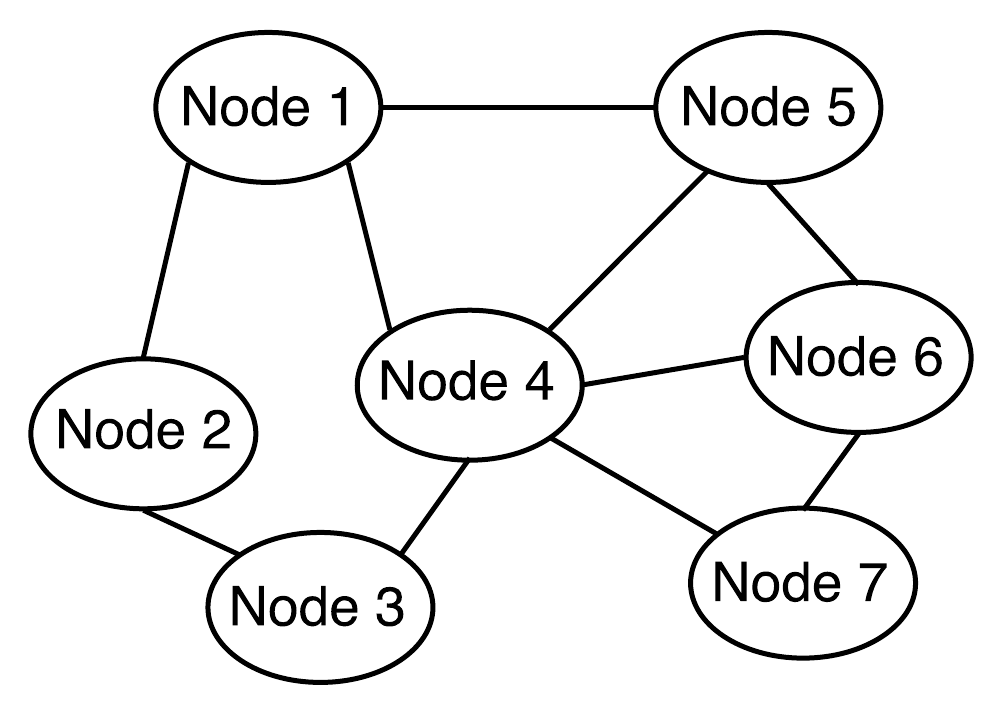}\label{fig:NetworkedSystem}}
\caption{Two Computing Structures to Perform SGD}\label{fig:TwoStructures}
\end{figure*}

In most of the current literature, the `distributed' SGD is usually performed by a 'one-Server multiple-Worker' mode, as shown in Fig.~\ref{fig:ServerWorker}, where the workers access the data distributedly and the server acts as a coordinator among all workers. In practice, the `server' and the workers can be many computers, connected by a computer network, like the design of the `parameter-server' \cite{li2014scaling}; also, the `server' can be the shared memory of one computer and the workers can be different threads, like the design of 'Hogwild!'\cite{recht2011hogwild}. To optimized the variable (training process), in each iteration, each worker will take some data samples to update their local variable and upload the information to the server; after receiving the new local variables from all workers, the server will perform a update, according to some requirements like budget limitation, and then broadcast the new variable to all workers. Intuitively speaking, the server exists so that all workers will be aware of the global information and the local variables of different workers will not deviate from each other too much.

However, the performance of the `server-worker' structure is inevitably limited by at least two factors. Firstly, due to the existence of `central controller', the system is actually semi-distributed. It is uneasy to scale up to too many workers. For example, it may need complicated design and advanced systems to handle tens of thousands of workers. Also, this kind of system is not so robust. For example, if the server breaks down, the entire system also fails to work. Secondly, synchronization among all workers is needed in the process. However, synchronization is difficult to achieve considering the heterogeneous and time-varying performances of different workers, especially when the number of workers is large. In practice, the late workers are simply ignored, which is equivalent to introducing noise and performance is thus degraded. In a recent work \cite{lian2015asynchronous}, the authors show that, for the `server-worker' structure, an asynchronized dual-based algorithm can achieve $O(1/\sqrt{T})$ convergence rate.

In this paper, we propose a SGD scheme for a general networked system, in which many nodes are connected by a general undirected graph, as shown in Fig.~\ref{fig:NetworkedSystem}. In this system, each node can represent a computing unit, like one computer of a cluster, one user of online social network, or one sensor of wireless sensor network. We assume that each node can access the data and do some local computation, but communications can only happens between directly connected nodes (the node and its neighbors). We design an update method without relying on any controller to achieve global optimality and consensus. More specifically, the local variables stored by each node will asymptotically converge to the same value, which is the optimal value to minimize the total expected loss of all nodes.

In particular, we make the following contributions.
\begin{itemize}
\item We reformulate the single-variable data-fitting problem of a networked system as a multiple-variable problem with multiple constraints, which ensure global consensus.
\item By apply the result of the recent paper on stochastic optimization problem with many constraints \cite{wang2015random}, we design an algorithm for the reformulated optimization problem. The algorithm is essentially distributed and asynchronized. Both the global consensus and optimality can be asymptotically achieved.
\item We analyze the convergence speed of algorithm for a special structure: the regular graph. Our analysis indicates a not so surprising result: the algorithm will converge faster for a graph with better connectivity.
\item By simulations on synthetic and real-world datasets, we validate the merit of our design and test the impact of different system settings.
\end{itemize}

\section{Problem Description}

We consider a scenario in which $N$ computing nodes are connected as a graph. We have one optimization variable $\mathbf{\beta}$ as a vector, and the loss of Node $i$ by $\mathbf{\beta}$ is denoted as $\bar{f}_i(\mathbf{\beta})$. One example is that Node $i$ has many training data samples, and $\bar{f}_i(\mathbf{\beta}) = \frac{1}{K_i}\sum_kl_i(\mathbf{\beta}, v_k)$ is the averaged loss of each sample $v_k$. We assume that $\bar{f}_i(\cdot)$ is a convex function. With one data sample being $v_k = (x_k,y_k)$, such a model can capture many popular examples as special cases, like
\begin{itemize}
\item Logistic Regression: $$\bar{f}_i(\mathbf{\beta}) = \frac{1}{K_i}\sum_{k=1}^{K_i}\left[y_k\frac{1}{1+e^{-\mathbf{\beta}^Tx_k}} + (1-y_k)\frac{e^{-\mathbf{\beta}^Tx_k}}{1+e^{-\mathbf{\beta}^Tx_k}}\right].$$
\item Support Vector Machine: $$\bar{f}_i(\mathbf{\beta}) = \frac{1}{K_i}\sum_{k=1}^{K_i}\max\{0,1-y_k\mathbf{\beta}^Tx_k\}+\lambda \mathbf{\mathbf{\beta}^2}.$$
\item Lasso: $$\bar{f}_i(\mathbf{\beta}) = \frac{1}{2K_i}\sum_{k=1}^{K_i}(y_k-\mathbf{\beta}^Tx_k)^2 + \lambda \|\mathbf{\beta}\|.$$
\end{itemize}

Our interest is to find one optimal solution to minimize the total loss of all nodes, \textit{i.e.}, solving
\begin{equation}\label{eq:problem_1}
\min\quad \frac{1}{N}\sum_{i=1}^N\bar{f}_i(\mathbf{\beta}).
\end{equation}

In this paper, instead of using the empirical loss, we consider minimizing a expected loss function $f_i(\beta) = \mathbb{E}\left[l_i(\mathbf{\beta}, V_i)\right]$, where the expectation is taken according to the distribution of $V_i$ \footnote{Note that we can assume that $V_i$ and $V_j$ can follow different distributions if $i\neq j$.}. $V_i$ is the random variable to generate a data sample in Node $i$. So our optimization problem is further casted as as a stochastic programming in the following form,
\begin{equation}\label{eq:problem_2}
\min\quad F(\beta) = \frac{1}{N} \sum_if_i(\beta).
\end{equation}

\textbf{Remarks}: (1) $\bar{f}_{i}(\mathbf{\beta})$ can be viewed as an empirical replacement of $f_{i}(\mathbf{\beta})$. By law of large numbers, the difference between two functions will converge to 0 if the number of data samples in Node $i$ goes to infinity. (2) Suppose we have one random variable $V$ to generate all data samples. We can have another perspective to interpret the objective function $F(\mathbf{\beta})$ of Problem.~\eqref{eq:problem_2}: $V$ has probability $\frac{1}{N}$ to take the distribution of that of $V_i$, and at the same time, the loss function will take the form of $l_i(\cdot,\cdot)$.

To summarize, we have an optimization problem with a single variable $\mathbf{\beta}$ to minimize the expected loss by data generated or stored in different nodes. The stochastic programming problem with convex objective functions has been extensively studied in the literature, and many off-the-shelf algorithm can be used, see \cite{wang2015random} and the reference therein. However, most of works consider centralized design and are not applicable to the networked system. The results in \cite{agarwal2010distributed,nedich2015convergence,nedic2009distributed} also consider to minimize an objective function over a networked system, but the algorithms require all nodes to perform update in each slot and synchronization is needed for real-world implementation, which, however, is difficult to achieve. In the next section, we will present our design to tackle this issue.

%is that we have one random variable $V$ with probability $\frac{1}{N}$ to take the distribution $V_i$ and the loss function has probability $\frac{1}{N}$ to take the form of $f_i(\cdot)$. To take a sample of

\section{Algorithm Design}

\subsection{Problem Reformulation}
Before presenting our algorithm, we denote $\mathcal{N}_i$ as the set of the neighbors of Node $i$ and reformulate original problem.~\eqref{eq:problem_2} as follows,
\begin{eqnarray}
\min_{\mathbf{\beta}_1,\mathbf{\beta}_2,\cdots, \mathbf{\beta}_N} && \quad\frac{1}{N}\sum_{i=1}^Nf_i(\mathbf{\beta}_i)\\
\text{s.t.}&&\quad \mathbf{\beta}_i = \mathbf{\beta}_k, \forall k \in \mathcal{N}_i, \forall i. \label{eq:consensus}
\end{eqnarray}

In the new formulation, we increase the number of variables by assigning one local optimization variable to each node. We assume that this networked system is connected, so for any feasible solution, global consensus is guaranteed by Eq.~\eqref{eq:consensus}. This reformulation is trivial at the first glance, but it is really helpful for the algorithm design and analysis, as we will show immediately.

\subsection{A Distributed Algorithm}

In \cite{wang2015random}, the authors investigated a general stochastic programming problem in the following form,
\begin{eqnarray*}
\textbf{GenPro}\quad\min_X &&\quad \mathbb{E}[F(X)]\\
\text{s.t.}&&\quad X \in \mathcal{X}_1\cap\mathcal{X}_2\cap\cdots\cap\mathcal{X}_M,
\end{eqnarray*}
and proposed several algorithms based on randomly sampling and projection. We recall one of them below,

\begin{algorithm}[H]
\protect\caption{An algorithm to solve \textbf{GenPro}}
\begin{algorithmic}[1]\label{alg:generalAlg}
\REQUIRE An oracle to generate data sample, $k=1$, $X^0$, stepsize $\alpha^t, t = 1,2,\cdots, \infty$
\ENSURE An optimal solution $X^*$ to \textbf{GenPro}
\WHILE{not converge}
\STATE Generate one data sample $v^k$ and compute the subgradient $g(X^k, v^k)$ with current variable $X^k$ and data sample $v^k$.
\STATE Update current variable $X^{k+1} = X^k - \alpha^k g(X^k, v^k)$.
\STATE Randomly pick one constraint $\mathcal{X}_m$, and make the projection $X^{k+1} = \Pi_{\mathcal{X}_m}(X^{k+1})$.
\STATE $k = k+1$
\ENDWHILE
\STATE $X^* = X^k$
\RETURN $X^*$.
\end{algorithmic}
\end{algorithm}

In Alg.~\ref{alg:generalAlg}, $\Pi_{\mathcal{X}_m}(\cdot)$ is a projection operator and $\Pi_{\mathcal{X}_m}(x) = \arg\min_{y\in\mathcal{X}_m}\|y-x\|_F$. Even though only simple gradient descent and projection are performed in each iteration, it can be proved that the algorithm will converge to a optimal solution, \textit{i.e.}, the objective value will converge to the optimal value and the distance from the variable to the feasible region will converge to 0.

We denote the optimization variable $\beta$ as a concentration of original local variables, \textit{i.e.}, $$\beta = \left[\beta_1,\beta_2,\cdots,\beta_N\right]$$ and $\mathcal{B}_n = \{\beta|\beta_{n} = \beta_k, \forall k \in \mathcal{N}_n\}$ as one convex set. The optimization problem we are interested in can be written in a more compact fact as
\begin{eqnarray}
\textbf{OurPro}\quad\min_{\mathbf{\beta}} &&\quad \quad\frac{1}{N}\sum_{i=1}^Nf_i(\mathbf{\beta}_i) \nonumber\\
\text{s.t.}&&\quad \beta \in \mathcal{B}_1\cap\mathcal{B}_2\cap\cdots\cap\mathcal{B}_N. \label{eq:consensus_0}
\end{eqnarray}

Now our problem are put into a form that is suitable for Alg.~\ref{alg:generalAlg}, so we can directly apply the algorithm. Now we will examine the gradient descent and projection steps in details.

\paragraph{Gradient Descent} To randomly generate a data sample, we can firstly randomly select a node $i$ and generate a sample $v_i^k$ according to the distribution of $V_i$. Note that under the condition node $i$ is selected, the loss of other nodes with $v_i^k$ is zero. So the sampled subgradient of the objective function is nonzero only for $\beta_i$ and can be written as $\frac{1}{N}\partial f_i(\beta_i)/\partial \beta_i$. Thus the gradient descent can be performed by
\begin{equation}\label{eq:gradientDescent}
\begin{cases}
\beta_i^{k+1} = \beta_i^{k} - \alpha^k \frac{1}{N}\partial l_i(\beta_i^k, v_i^k)/\partial \beta_i^k,\\
\beta_n^{k+1} = \beta_n^k, \quad\text{for } n \neq i.
\end{cases}
\end{equation}

We can see that if a data sample is generated locally in Node $i$, only \textbf{local update} is needed to perform gradient descent.

\paragraph{Projection} If one convex set, say $\mathcal{B}_m$, is randomly selected, we would need to project the current variable $\left[\beta_1^{k+1}, \beta_2^{k+1},\cdots,\beta_{N}^{k+1}\right]$ onto it. $\mathcal{B}_m$ requires that the local variables stored in node $m$ and its neighbors are equal. To find a new point satisfy this condition and minimize the distance between the new point and original point, the values of the variables stored in node $m$ and its neighbors should take the averaged value among them and the values of other variables should stay the same. Thus the projection can be performed by
\begin{equation}\label{eq:Projection}
\begin{cases}
\beta_i^{k+1} = \frac{1}{1+|\mathcal{N}_m|}\sum_{l \in \{m\}\cup\mathcal{N}_m} \beta_l^k, \quad \forall i \in \{m\}\cup\mathcal{N}_m,\\
\beta_j^{k+1} = \beta_j^k,\quad \text{for other node $j$}.
\end{cases}
\end{equation}

We can also see that the projection process only involves one node and its neighbors. Specifically, one node collects the local information from its neighbors, calculates the average and then makes a broadcast to its neighbors.

We summarize the above procedure to solve \textbf{OurPro} in Alg.~\ref{alg:ourAlg} to end this part.

\begin{algorithm}
\protect\caption{An algorithm to solve \textbf{OurPro}}
\begin{algorithmic}[1]\label{alg:ourAlg}
\REQUIRE An oracle to generate data sample, $k=1$, $\beta^0$, stepsize $\alpha^t, t = 1,2,\cdots, \infty$
\ENSURE An optimal solution $X^*$ to \textbf{OurPro}
\WHILE{not converge}
\STATE Randomly select a node $i$, and then generate a random variable $r$ from 0 to 1.
\IF{$r \leq 0.5$}
\STATE Generate one data sample $v_m^k$ and update the local variable by $\beta_m^{k+1} = \beta_m^{k} - \alpha^k \frac{1}{N}\partial l_m(\beta_m^k, v_m^k)/\partial \beta_m^k.$
\ELSE
\STATE Collect information from its neighbors and make a broadcast so that $\beta_i^{k+1} = \frac{1}{1+|\mathcal{N}_m|}\sum_{l \in \{m\}\cup\mathcal{N}_m} \beta_l^k, \quad \forall i \in \{m\}\cup\mathcal{N}_m$
\ENDIF
\STATE $k = k+1$
\ENDWHILE
\STATE $\beta^* = \beta^k$
\RETURN $\beta^*$.
\end{algorithmic}
\end{algorithm}

\subsection{Theoretical Results}

In this part, we present some theoretical results related to the feasibility and optimality of Alg.~\ref{alg:ourAlg}. Unless otherwise mentioned, we will assume that the objective function satisfies Assumption 1 in \cite{wang2015random} and stepsizes are set properly. For a concise presentation, we use $DF(\beta^k)$ as the distance from the current optimization variable to the feasible region $\mathcal{B}_1\cap\mathcal{B}_2\cap\cdots\cap\mathcal{B}_N$ \footnote{It should be noted that this region is a polyhedron and can be characterized by linear constraints.} and $DO(\beta^k)$ as the distance from the current optimization variable to the optimal region.

\begin{theorem}\label{theorem:feasibilityandoptimality}
{(\text{Feasibility} and \text{Optimality}) \cite{wang2015random}}
 Both $DF(\beta^k)$ and $DO(\beta^k)$ will converge to zero almost surely.
\end{theorem}

We also provide the asymptotic convergence speed in the following theorem.
\begin{theorem}\label{theorem:convergencespeed}
{(\text{Convergence Speed})\cite{wang2015random}} We can have
\begin{align}
\mathbb{E}[DF(\beta^{k+1})] \leq (1-\frac{C}{4})DF(\beta^k) + \sigma(5+\frac{4}{C})\alpha_k^2, \label{eq:convergencespeed_DF}
\end{align}
and
\begin{align}
&\mathbb{E}[DO(\beta^{k+1})] \nonumber\\
\leq & DO(\beta^k) + \sigma(5+\frac{2}{C})\alpha_k^2 - 2\alpha_k(F(\Pi(\beta^k)- F^*), \label{eq:convergencespeed_DO}
\end{align}
where $\sigma$ is a constant determined by the objective function, $\alpha_k$ is the stepsize, and $C = \frac{\eta}{N}$ ($\eta$ is the value for the linear regularity condition for the feasible region \eqref{eq:consensus_0}. \footnote{The definition of `linear regularity condition' is provided in the appendix. }).
\end{theorem}

We omit the proofs of Theorem~\ref{theorem:feasibilityandoptimality} and \ref{theorem:convergencespeed} since they are established directly from the results in \cite{wang2015random} (on Page 9 and Page 10), but we want to provide some results specifically related to our problem. Note from Theorem~\ref{theorem:convergencespeed} that a larger value of $C$ will lead to faster convergence. On the one hand, $C$ will be larger if this network contains less nodes ($N$ is smaller), which is intuitive. On the other hand, a larger value $\eta$ will also help the algorithm to converge faster. We provide some results for more insights below.

Firstly, for a general graph, we define a matrix for local averaging $A = \left[a_{i,j}\right]$ where $a_{i,j} = \frac{1}{1+|\mathcal{N}_i|}$. The meaning of $A$ can be explained as it computes the new value for one node as the average of the original value of itself and its neighbors.

\begin{lemma}\label{LEMMA:ETA}
For a system connected by a \textbf{k-regular} graph, a lower bound for the value of $\eta$ in Theorem.~\ref{theorem:convergencespeed} is given by
$$(1-\sigma_2^2)\frac{k+1}{N},$$
where $\sigma_2$ is a second largest eigenvalue of its averaging matrix $A$.
\end{lemma}

The proof is deferred in Appendix.~\ref{PROOF:LEMMA:ERA} and we provide some remarks to end this part.

\textbf{Remarks}: (a) Both a larger value of $k$ and a smaller value of $N$ will increase the value of $\eta$ and thus increase the convergence speed. This phenomenon can be intuitively understood because the algorithm will converge faster for a better-connected and smaller-scale system. (b) If $\sigma_2$ is smaller, the algorithm will also converge faster. Essentially, the second largest singular value of $A$ is another metric to evaluate the `connectivity' of the system. Remember that the mixing rate of the Markov chain is also affected by the second largest eigenvalue of the transition probability matrix. The readers can refer to \cite{boyd2004fastest} for more information.

\section{Discussions of Practical Implementations}

In this part, we try to discuss some implementation details of Alg.~\ref{alg:ourAlg}.

\subsection{Node Selection}

In Alg.~\ref{alg:ourAlg}, one node is randomly selected in each iteration, to perform gradient decent or local average. One trivial way to do this is to generate a random integer form 1 to $N$, and evoke the corresponding node according to this results. However, this method cannot be realized without a central controller. A distributed alternative would be that, we let each node generate a random variable according to a geometric distribution and then count backwards. It will be `selected' when it counts to 0. We can even carefully design the parameter of geometric distributions for different nodes so that the probability for different nodes to be selected is preferred. The rationale behind this mechanism is similar to that of Monto Carlo Markov Chain Sampling \cite{gilks2005markov}, CSMA \cite{chen2013markov}, \textit{etc}.

\subsection{Communication Overhead}

For the local average update, communication is needed between the selected node and its neighbors. A common understanding is that communication between nodes are much less efficient than local computation and is unpreferred. To decrease the communication overhead, we can decrease the probability to perform local average when one node is selected. But this mechanism will decrease the convergence speed to global consensus.

\subsection{Update Conflict }

Since we select the nodes in a distributed manner, it is possible that two nodes are selected in the same time slot. (Imagine that two nodes generate the same value and count down to 0 at the same time.) If the two nodes are far away from each other, \textit{i.e.}, they share no neighbors, updating simultaneously does no harm and is even preferred since it is equivalent to do the update one by one. However, if two nodes are connected, such updates will introduce some conflicts, say, one node plans to do gradient descent but its neighbor tells him to update according to average. One proper way to handle this issue is that, if one node is `selected', it sends a message to its neighbors for locking up. Such an mechanism can avoid update conflict but introduce additional communication overhead.

\section{Simulations}

In this part, we try to evaluate the performance of Alg.~\ref{alg:ourAlg} and the impact of different parameters. We describe the simulation setting in Sec.~\ref{sec:simulationsettings}. The results from Sec.~\ref{sec:syntheticdata0} to Sec.~\ref{sec:syntheticdata1} are based on synthetic data while the results in Sec.~\ref{sec:realdata} are based on real-world data.

\subsection{Simulation Data and Settings}\label{sec:simulationsettings}
In our simulation, we perform a multiple class classification task by logistic regression (multinomial logistic regression). The objective of the training process is to minimize the cross entropy between the empirical distribution and predicted distribution, which is a convex function. From Sec.~\ref{sec:syntheticdata0} to Sec.~\ref{sec:syntheticdata1}, we let each node have its own distribution to generate data sample. We have 10 categories and 50 features. Also, we assume that the distributions for different nodes are different, so training with only one or several nodes will deviate from the global optimality. In Sec.~\ref{sec:realdata}, we test the performance by a real-world dataset: \textbf{notMNIST}, the size of which is around 12G. For this dataset, we have 10 categories and 256 features. 

All simulations are implemented by tensorflow, and the codes are available \cite{code}.

\subsection{Global Consensus}\label{sec:syntheticdata0}
\begin{figure}
  \centering
  \includegraphics[width=0.95\columnwidth]{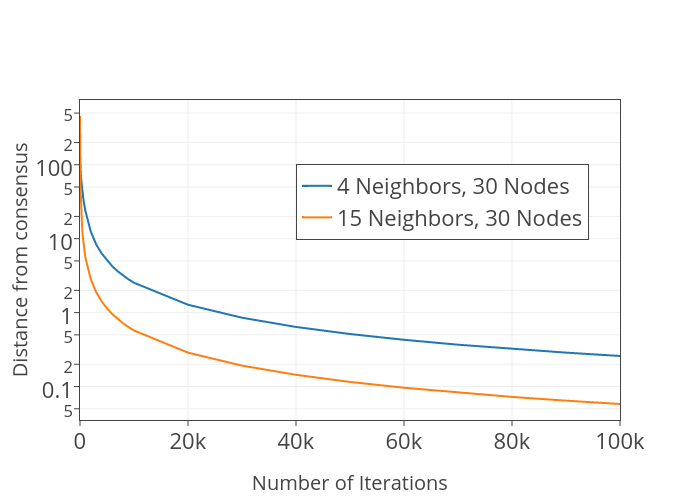}\\
  \caption{Distance to global consensus}\label{fig:sim_consensus}
\end{figure}

Our first interest about Alg.~\ref{alg:ourAlg} is to show whether it can guarantee global consensus and how much time it takes to achieve consensus. Towards this end, we define $d^k = \sum_{i=1}^N\|\beta_i^k-\bar{\beta}^k\|$ as the `distance of the variables from global consensus' in time slot $k$, where $\bar{\beta}^k = \frac{\sum_{i=1}^N\beta_i^k}{N}$. We test the performance of two 30-node systems connected by a regular graph: one by 4-regular graph and the other one by 15-regular graph. The result is depicted in Fig.~\ref{fig:sim_consensus}.

Note that the y axis is logarithmic, and we can see that $d^k$ converges very fast to zero. More specifically, the value of $d^k$ is below 10 after 10k updates. Considering the number of the features (50) and nodes (30), this result indicates that the global consensus is almost achieved. Another observation is that it converges faster for the 15-regular graph, which fits the result of Lemma.~\ref{LEMMA:ETA} and our intuition. 

\subsection{Prediction Error}\label{sec:predictionerror}
\begin{figure}
  \centering
  \includegraphics[width=0.95\columnwidth]{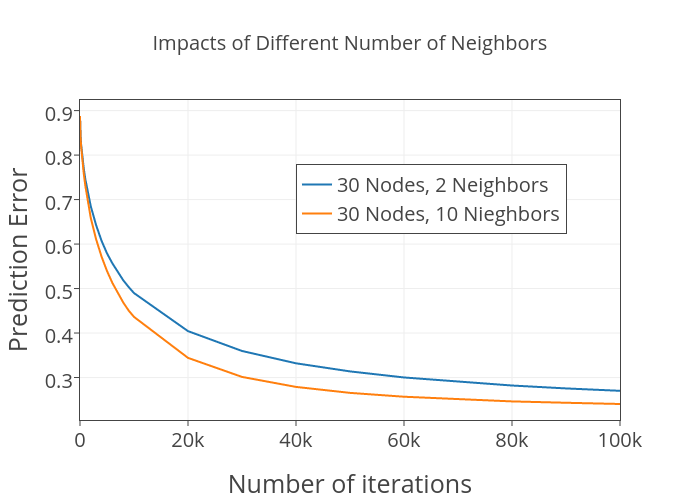}\\
  \caption{Prediction error}\label{fig:sim_optimality}
\end{figure}

We also test the prediction error by $\bar{\beta}^k$, the averaged value of current variables on all nodes. We test two 30-node systems: one by 2-regular graph and the other one by 10-regular graph. The results are reported in Fig.~\ref{fig:sim_optimality}.

As we can see, in both cases, the prediction error decreases with more iterations (more data feeding). The prediction errors will be under 0.4 after 40k iterations (a random guess will lead to 0.9 prediction error) and the prediction error decreases faster for the 10-regular graph. Considering the facts that we add noise to the generated data samples in training process and the distributions of different nodes are different, we can conclude that our mechanism is very effective. 

\subsection{With Network Size Increasing}\label{sec:syntheticdata1}

\begin{figure}
  \centering
  \includegraphics[width=0.95\columnwidth]{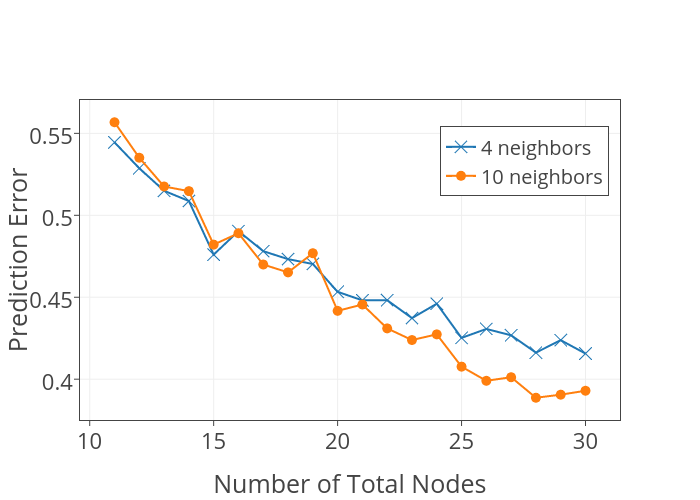}\\
  \caption{With more nodes joining }\label{fig:morenodes}
\end{figure}

One motivation of this work is to build a scalable and robust system to enjoy the merit of `big data'. It is necessary to ask whether we can we predict more accurately when more nodes join the system? To answer this question, we test the final prediction error with the number of nodes increasing from 10 to 30, and each node generates 500 data samples on average. Also we set the number of neighbors for each node to be 4 and 10 respectively. The result is shown in Fig.~\ref{fig:morenodes}.

As we can see from Fig.~\ref{fig:morenodes}, the decreasing trend of the prediction error with more nodes is quite clear (not always because the stochastic nature of the algorithm). Also, the advantage of a better connected system is more clear for a larger system (the number of nodes is larger). The rationale behind this phenomenon can be explained as follows. When the size of the system is small, the distances between each two nodes are relatively small, so the information of one node can easily spread out to all nodes, even though the number of neighbors of each node is small. 

\subsection{Simulation based on a real-world dataset}\label{sec:realdata}

\begin{figure}
  \centering
  \includegraphics[width=0.95\columnwidth]{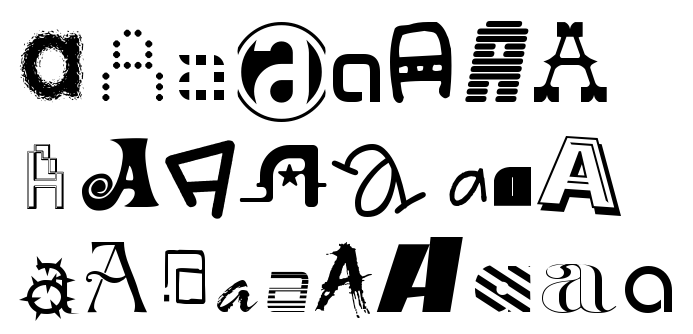}\\
  \caption{Letter `A' in notMNIST \cite{notmnist}}\label{fig:notmnist}
\end{figure}

We also test the performance of Alg.~\ref{alg:ourAlg} on a real-world dataset \textbf{notMNIST}. This dataset contains the images of characters in the alphabet and digits. A glace of the letter `A' in the dataset is shown in Fig.~\ref{fig:notmnist}.

\begin{figure}
  \centering
  \includegraphics[width=0.95\columnwidth]{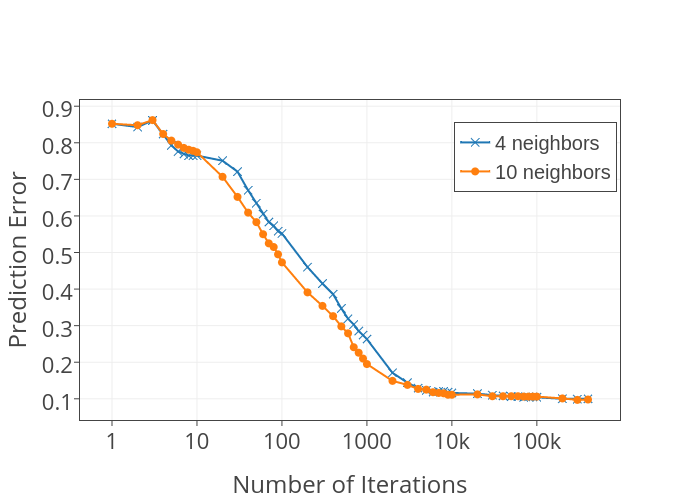}\\
  \caption{Prediction error(notMNIST)}\label{fig:sim_optimality_real_world}
\end{figure}

We select 10 categories to form a dataset to perform a multinomial logistic regression task. The simulation setting is similar to that of Sec.~\ref{sec:predictionerror}. And we report the result in Fig.~\ref{fig:sim_optimality_real_world}.

As we can see, the prediction error converges to less than 0.1, which is almost the same result of a centralized version of SGD. Also, the performances of two systems with different connectivity converge to the same value. This result indicates that as long as the system is connected, optimality will be asymptotically achieved, even the convergence speed will be affected by the system topology.  

\section{Conclusion}
In this paper, we propose a fully distributed and asynchronized stochastic gradient method for a networked system. We augment the variable size and reformulate a stochastic programming problem with many constraints. By adapting some results in recent literature, we can achieve global consensus and optimality by only local computation and communication. A sharp lower bound of the convergence speed for the regular graph case is characterized and some simulation results are also provided to validate the merit of our design. One possible future work is to implement this algorithm on real-world system, especially on the heterogeneous system including high-performance computing clusters and low-performance mobile devices.

\bibliographystyle{abbrv}
\bibliography{ref}

\appendix
\section{Linear Regularity Condition}\label{sec:proof_prop_relax}
\subsection{Definition}

We say that the linear regularity condition holds for $\mathcal{B} = \mathcal{B}_1\cap\mathcal{B}_2\cap\cdots\cap\mathcal{B}_N$ with a constant scalar $\eta\in(0,1)$ if for all $x$ in the space, we can have
$$\eta\|x-\Pi_{\mathcal{B}}(x)\|^2 \leq \max_{i = 1,\cdots, N}\|x-\Pi_{\mathcal{B}_i}(x)\|^2.$$

Some explanations are provided below for a better understanding.

\begin{itemize}
\item If the linear regularity condition hold for $\eta_0$, then it also holds with $\bar{\eta} \in (0, \eta_0)$.
\item For the result in Theorem.~\ref{theorem:convergencespeed}, a larger $\eta$ will lead to faster convergence speed.
\item For the constraints satisfy the linear regularity condition, projecting onto them sequentially with many rounds (a cyclic projection method) will converge to their intersection.
\item The linear regularity condition automatically hold when $\mathcal{B}_i$ are linear half spaces \cite{deutsch2008rate}. So it holds for Eq.~\eqref{eq:consensus_0}.
\end{itemize}

\subsection{Proof of Lemma.~\ref{LEMMA:ETA}}\label{PROOF:LEMMA:ERA}

\begin{proof}
Recall that $\beta = [\beta_1,\beta_2,\cdots, \beta_N]$. Let us suppose we have a random variable $Y$, uniformly distributed on $\{\beta_1,\beta_2, \cdots, \beta_N\}$. Besides, we have another random variable $I$ distributed on $\{1,2,\cdots,N\}$. Conditioning on that $I = i$, $Y$ will be uniformly distributed on $\{\beta_k| k\in \{i\}\cup\mathcal{N}_i\}$.

By law of total variance, we can have
$$\text{Var}(Y) = \text{E}\left[\text{Var}(Y|I)\right] + \text{Var}(\text{E}[Y|I]).$$

Note that $$\text{Var}(Y) = \frac{1}{N}\|\beta-\Pi_{\mathcal{B}}(\beta)\|$$ and
\begin{align*}
&\text{E}\left[\text{Var}(Y|I)\right] \\
= &\sum_{i}P_i\frac{1}{1+|\mathcal{N}_i|}\|\beta-\Pi_{\mathcal{B}_i}(\beta)\|^2\\
\leq & \sum_{i}P_i \max_i \frac{1}{1+|\mathcal{N}_i|} \|\beta-\Pi_{\mathcal{B}_i}(\beta)\|^2\\
\leq &\frac{1}{1+k}\max_i \|\beta-\Pi_{\mathcal{B}_i}(\beta)\|^2 (\text{for a k-regular graph})
\end{align*}

If we can prove that
\begin{equation}\label{eq:critical}
\text{Var}(\text{E}[Y|I]) \leq \sigma_2^2\text{Var}(Y),
\end{equation}
we can have
$$\text{Var}(Y) \leq  \frac{1}{1+k}\max_i \|\beta-\Pi_{\mathcal{B}_i}(\beta)\|^2 + \sigma_2^2\text{Var}(Y).$$

Then
\begin{align*}
&(1-\sigma_2^2)\text{Var}(Y) = (1-\sigma_2^2) \frac{1}{N}\|\beta-\Pi_{\mathcal{B}}(\beta)\|\\
\leq &  \frac{1}{1+k}\max_i \|\beta-\Pi_{\mathcal{B}_i}(\beta)\|^2,
\end{align*}
which is the result of Lemma.~\ref{LEMMA:ETA}.

To complete the proof, the remaining part is to show Eq.~\eqref{eq:critical}.

We define another matrix $\bar{A} = [\frac{1}{N}]_{N\times N}$.
Note that
\begin{align*}
&\text{Var}(Y)\\
= &\frac{1}{N}\|(I-\bar{A})\beta\|_2 \\
= &\frac{1}{N}\beta^T(I-\bar{A})^T(I-\bar{A})\beta,
\end{align*}
and
\begin{align*}
&\text{Var}(\text{E}[Y|I])\\
= & \frac{1}{N}\|(A-\bar{A})\beta\|_2\\
= & \frac{1}{N}\beta^T(A-\bar{A})^T(A-\bar{A})\beta.
\end{align*}

Now it would be enough to show that $\sigma_2^2 (I-\bar{A})^T(I-\bar{A}) - (A-\bar{A})^T(A-\bar{A})$ is positive semidefinite.
This statement holds with the following facts.
\begin{itemize}
\item $A$ is a doubly stochastic matrix. Its largest singular value is 1, with singular vector being $\left[\frac{1}{N},\frac{1}{N},\cdots, \frac{1}{N}\right]$. We denote its eigenvalue decomposition by $A = V\Sigma V^T$, where $\Sigma$ is a diagonal matrix.
\item $\bar{A}$ is of rank 1 and it can be decomposed as $\bar{A} = V\text{Ind}V^T$, where $\text{Ind}_{i,j} = \begin{cases}1 \text{ if } i = j = 1, \\0 \text{otherwise} \end{cases}$.
\end{itemize}

The proof is completed.

\end{proof}

\iffalse

\input{texFile/Introduction.tex}
\input{texFile/SystemModelProblemFormulation.tex}

\input{texFile/AlgorithmDesign.tex}

\input{texFile/Simulation.tex}

\input{texFile/Conclusion.tex}

%\input{Remark.tex}
\bibliographystyle{abbrv}
\bibliography{ref}

\input{texFile/Appendix.tex}
\fi
\end{document}